\documentstyle[11pt,a4,psfig] {article}

\title{Integrated speech and morphological processing in a connectionist
continuous speech understanding for Korean}

\author{Geunbae Lee \\ Jong-Hyeok Lee \\ 
Department of Computer Science \& Engineering \\
Pohang University of Science \& Technology \\
San 31, Hoja-Dong, Pohang, 790-784, Korea \\
Phone: +82-562-279-2254, Fax: +82-562-279-2299 \\
gblee@vision.postech.ac.kr, jhlee@vision.postech.ac.kr 
}

\date{}

\begin{document}

\maketitle

\begin{abstract}
A new tightly coupled speech and natural language integration
model is presented for a TDNN-based continuous possibly large vocabulary
speech recognition system for Korean. Unlike popular n-best techniques 
developed for integrating mainly HMM-based speech recognition and natural 
language processing in a {\em word level}, which is obviously inadequate for 
morphologically complex agglutinative languages, our model constructs a spoken
language system based on a {\em morpheme-level} speech and language 
integration. With this integration scheme, the
spoken Korean processing engine (SKOPE) is designed and
implemented using a TDNN-based diphone recognition module
integrated with a Viterbi-based lexical decoding and symbolic 
phonological/morphological co-analysis. 
Our experiment results show that the speaker-dependent continuous 
{\em eojeol} (Korean word) recognition and integrated morphological analysis 
can be achieved with over 80.6\% success rate directly from speech inputs for 
the middle-level vocabularies. \\

\end{abstract}

\section{Introduction}
\begin{figure}
\centerline{\psfig{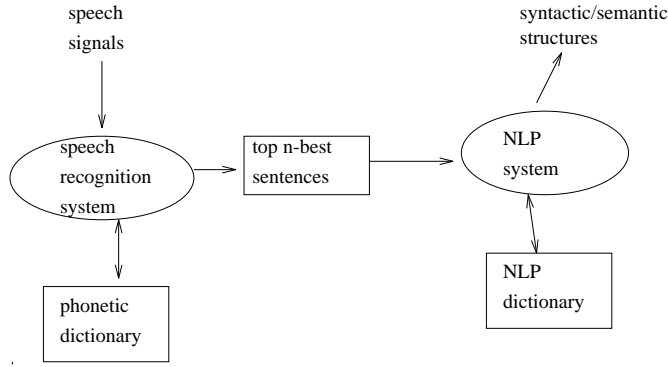}}
\caption{N-best lists: current speech and natural language 
integration method} \label{fg:nbest}
\end{figure}

A spoken natural language system requires many different levels
of knowledge sources including acoustic-phonetic, phonological, morphological,
syntactic, semantic and even pragmatic levels. These knowledge sources are grouped
and processed by either speech processing models or statistical/symbolic 
natural language processing models. Since the speech and the natural language
communities have conducted almost independent researches, these
models were not completely integrated and often biased by
neglecting either acoustic-phonetic or high-level linguistic
information. 
Current speech and natural language integration mainly relies on 
word-level n-best search techniques \cite{chow:nbest,schwartz:efficient} as
shown in figure~\ref{fg:nbest}.  
For HMM-based speech recognition systems, the n-best search techniques 
have been successfully applied to the integration of 
speech and natural language processing. 
However, current implementations of n-best techniques
only support the integration at a word level by directly producing the n-best list of
candidate sentences, and this type of loose coupling is only suitable for the 
integration of existing speech and natural language systems, such as, e.g.
\cite{agnas:spoken,bates:bbn}. The n-best search is viable only for
short sentences since the necessary n grows exponentially with the sentence
length. Because the n-best search directly generates word sequences,
phonetic and natural language dictionaries must have full 
word entries, which is obviously inadequate for morphologically complex 
agglutinative languages such as Korean. The dictionary size 
will grow very fast for full word entries because new words can be
almost freely generated by concatenating the constituent morphemes in these
languages (e.g. noun plus noun-endings or verb plus verb-endings). 

In this paper, we present a new morphologically conditioned integration architecture 
of speech
and natural language processing for morphologically complex agglutinative
languages. The integration is based on a Viterbi-based lexical
decoding and symbolic phonological/morphological co-analysis. 
The Viterbi search \cite{forney:viterbi} is performed on diphone (explained
in section~\ref{sec:diphone}) sequences generated from a 
TDNN (time-delay neural network)-based Korean speech recognizer
\cite{kim:integrating}, and the search process is tightly integrated with 
a morphological and phonological constraint checking. 
We present a new integration architecture, 
not for popular HMM-based systems, but for recently developed connectionist 
speech recognition systems.  Connectionist speech recognition 
has several advantages over the classical statistical 
speech processing \cite{morgan:neural}. Especially, the 
TDNN model \cite{waibel:phoneme} has been widely used to model
the time shift invariance of speech signals. In this regard, we will present a 
morpheme-level integration 
method for a TDNN-based continuous speech recognition model for Korean. 
This paper is organized as follows. Section~\ref{sec:features} briefly explains the
characteristics of spoken Korean for general readers. Section~\ref{sec:skope}
introduces our speech and natural language integration architecture,
and section~\ref{sec:diphone} and section~\ref{sec:viterbi} more elaborate the
introduced integration architecture. Section~\ref{sec:imple} shows several 
experiment results to
demonstrate the performance, and section~\ref{sec:comp} compares our integration
scheme with similar related researches. Section~\ref{sec:con} draws some
conclusions. 

\section{Features of spoken Korean}
\label{sec:features}
This section briefly explains the linguistic characterists of spoken
Korean before describing the integration architecture. In this paper, 
Yale romanization is used for representing the Korean phonemes.
1) A Korean word, called {\em eojeol}, consists of more than one morphemes
with clear-cut morpheme boundaries (Korean is an agglutinative language). 
2) Korean is a postpositional language with many kinds of noun-endings,
verb-endings, and
prefinal verb-endings. These functional morphemes determine the noun's
case roles, verb's tenses, modals, and modification relations between
eojeols. 
3) Korean is a basically SOV language but has relatively free word order 
compared to the rigid word-order languages, such
as English, except for the constraints that the verb must appear
in a sentence-final position. However, in Korean, some word-order 
constraints do exist
such that the auxiliary verbs representing modalities must follow the main
verb,
and the modifiers must be placed before the word (called head) they modify.
4) The unit of pause in speech (which is called {\em eonjeol}) may be
different
from that of a written text (an eojeol). The spoken morphological analysis 
must deal with 
an eonjeol (fragment of sentence) since no eojeol boundary is provided in 
the speech.  
5) Phonological changes can occur in a morpheme, between morphemes in an
eojeol, and even between eojeols in an eonjeol. These changes include consonant and
vowel assimilation, dissimilation, insertion, deletion, and contraction, and so on. 

\section{SKOPE system architecture for morpheme-level integration}
\label{sec:skope}
The morpheme-level integration technique processes phoneme-like 
unit (PLU) sequences (speech recognizer's outputs) using both
Viterbi-based lexical decoding (for morpheme) and symbolic 
phonological/morphological co-analysis, and uses a
single unified phonetic-morpheme (UPM) dictionary for both speech and 
language processing. This morpheme-level integration scheme is able to utilize
natural language morphological processing techniques in an
early stage of spoken language processing compared with the classical approaches 
of word-level speech and language integration. 
The morpheme-level integration also renders a phonological rule modeling possible
in the early stage. The phonological/morphological
analysis can be performed together using the single UPM
dictionary, and the dictionary size becomes stable regardless of
the vocabulary size because only the morphemes are encoded and the new words
can be processed by using the existing morphemes in the dictionary. 

\begin{figure}
\centerline{\psfig{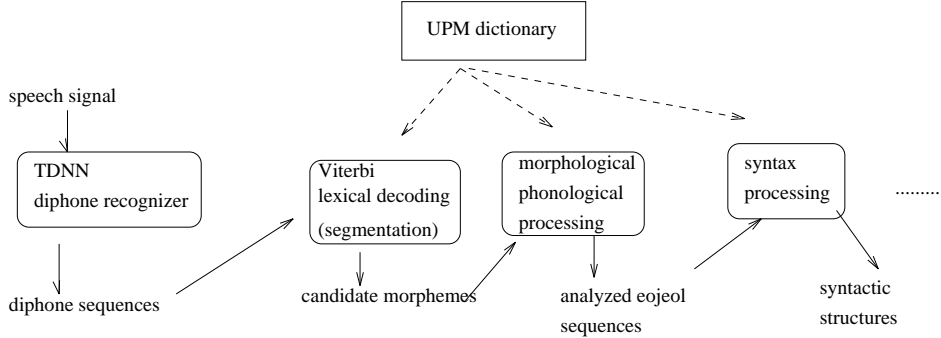}}
\caption{SKOPE speech and language morpheme-level 
integration architecture.
Syntax and more high-level processing steps are not in the scope of this
paper.} \label{fg:arch}
\end{figure}

Figure~\ref{fg:arch} shows the SKOPE architecture, a morpheme-level 
integration model of speech and natural language processing for Korean. 
The speech signal is analyzed using the TDNN diphone
recognizer. The diphone recognizer is composed of a hierarchy of
TDNN networks. The recognized diphone sequences are decoded using the Viterbi search
on the trie-structured UPM dictionary to segment out the
target morpheme candidates.  
In the UPM dictionary, each morpheme's phonetic header is a HMM (hidden markov model) 
network using the diphone symbols. The Viterbi decoded candidate morphemes are 
stored in a triangular table to be properly connected during the morphological 
processing. From the candidate morphemes, the 
Viterbi-based morphological analyzer produces the morphologically 
analyzed {\em eojeols} 
by handling morphotactics verification and irregular conjugations.
The phonological modeling is tightly integrated into the morphological
processing through a declarative phonological rule modeling in the UPM 
dictionary. Outputs of the integrated architecture, that is, analyzed
{\em eojeol} sequences, can be directly fed to the upper 
level syntax and semantics analysis modules which are described in \cite{lee:table}. 

\section{Diphone-based speech recognition}
\label{sec:diphone}
\begin{figure}
\centerline{\psfig{figure=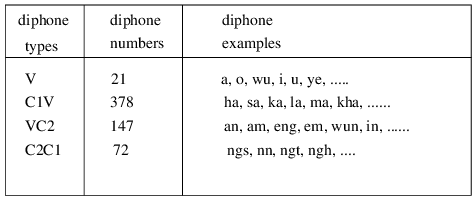}}
\caption{Korean diphone groups (V: vowel, C1: syllable-first
consonant, C2: syllable-final consonant). In C2C1 type, the C2 must be one
of the nasals or liquids/glides which are similar to vowels. 
Yale romanization is used to specify the diphone symbols.} \label{fg:diphone}
\end{figure}

\begin{figure}
\centerline{\psfig{figure=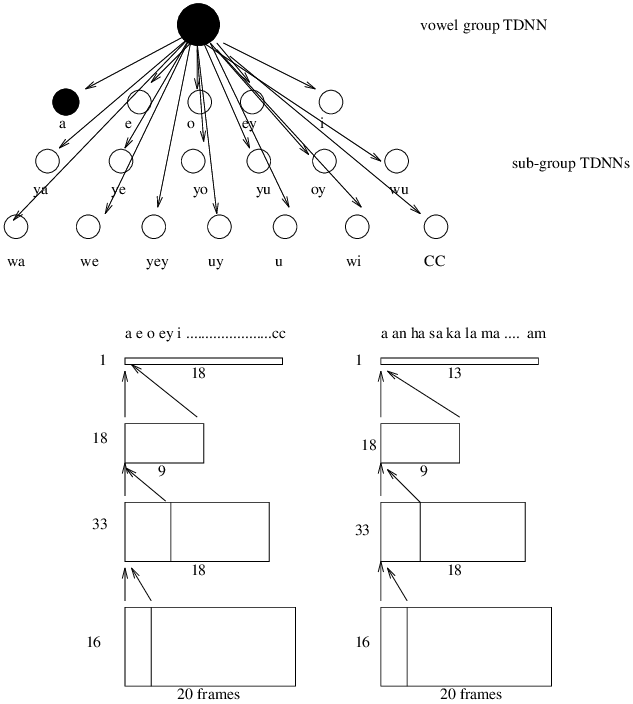}}
\caption{Top: hierarchical organization of the group of TDNNs for 
entire diphone recognition. Bottom left: TDNN architecture for vowel 
group identification. Note the cc group contains no vowels. Bottom 
right: Architecture of the sub-TDNN 
for /a/ vowel group recognition. The other 17 sub-TDNNs have the same 
architecture, but different number
of output units according to the number of diphones in each of the
vowel group.} \label{fg:dtdnn}
\end{figure}

For large-vocabulary continuous speech recognition, a sub-word
level recognition is usually performed. We select a group of
diphones for our phoneme-like units (PLUs) because direct phoneme recognition
in Korean is very difficult. The 46 Korean phonemes are very similar
each other especially in the following cases: 1) the Korean
diphthongs are hard to distinguish from the mono-vowels, and 2)
the syllable-final consonants are hard to differentiate from the
syllable-first consonants. The selected diphone groups 
(figure~\ref{fg:diphone})
have more suitable features for co-articulation modeling than the phonemes 
and are much fewer in numbers than the popular triphones \cite{lee:auto}.
We also introduced CC-type (syllable-final consonant, syllable-first 
consonant) diphones for smooth transition modeling between syllables in Korean.
Figure~\ref{fg:dtdnn} shows the hierarchical structure of a group of TDNNs
for diphone recognition, and also shows the architecture of each component TDNN. 
The whole diphone recognizer consists of total 19
different TDNNs for recognition of the defined Korean diphones.
We re-classified the total
diphones into 18 different groups according to the vowel characteristics 
in the diphones.
The top-level TDNN (vowel group TDNN) identifies the 18 vowel groups 
of the diphones using relatively low frequency signal vectors (under 4 KHz). 
Each 18 different sub-group TDNN recognizes the target diphones using the
whole frequency signal vectors.
For the training of each TDNN, we manually segmented the digitized speech signals
into 200 msec range (which includes roughly left-context phoneme, target
diphone, and right context phoneme), and applied 512 order FFTs and 16 step
mel-scaling \cite{waibel:phoneme} to get the filter-bank coefficients. Each
frame size is 10
msec, so 20 (frames) by 16 (mel-scaling factor) values are fed to
the TDNNs with the proper output symbols, that is, the vowel group name or 
the target diphone name.
After the training of each TDNN, the diphone recognition is performed by
feeding 200 msec signals to the vowel group TDNN
and subsequently to the
proper sub-group TDNNs according to the extracted vowel group. 
The 200 msec signals are shifted by 30 msec
steps and continuously fed to the networks to process the
continuous speech in an {\em eonjeol} (pause unit of Korean speech).  
The final outputs are sequence of diphones for each 200 msec range in 30
msec intervals. The hierarchical TDNN structure shortens the training time
and provides easily extensible system design. The entire recognition rate
critically depends on the vowel group TDNN in this hierarchical structure. 

\section{Viterbi-based morphological analysis}
\label{sec:viterbi}
Unlike conventional morphological analyses for text inputs, our 
morphological analysis starts with the recognized diphone sequences which
contain insertion, deletion, and substitution speech recognition errors. 
The conventional
morphological analysis procedure \cite{sproat:morph}, i.e., morpheme segmentation, 
morphotactics modeling, and orthographic rule (or phonological rule) modeling, 
must be augmented and extended to cope with the recognition errors as follows: 
1) The conventional morpheme segmentation is extended 
to deal with the speech recognition errors and 
between-morpheme phonological changes as well as irregular conjugations 
during the segmentation, 2) the
morphotactics modeling is extended to cope with the complex verb-endings and 
noun-endings in Korean, and 3) the orthographic rule modeling is combined with 
the phonological rule modeling to correctly transform the diphone
transcriptions (phonetic spelling) into the orthographically spelled morpheme 
sequences. 

\begin{figure}
\centerline{\psfig{figure=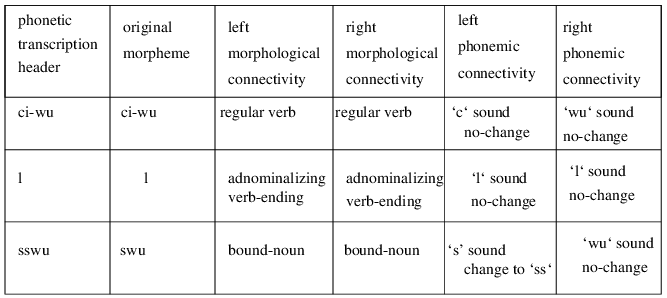}}
\caption{The unified phonetic-morpheme (UPM) dictionary for entries 
{\em ci-wu} (delete), {\em l} (adnominalizing verb-ending), and 
{\em swu} (bound-noun).} \label{fg:dict}    
\end{figure}

\begin{figure}
\centerline{\psfig{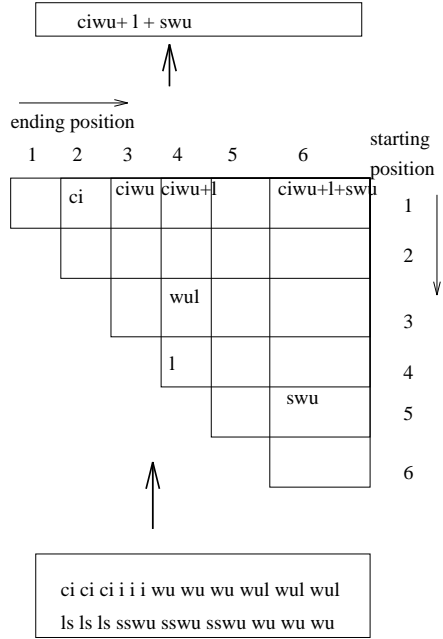}}
\caption{Morphological analysis of diphone sequences. From top:
output morpheme sequence in an {\em eonjeol}, triangular parsing table, and input 
diphone sequence.} 
\label{fg:morph}
\end{figure}

\begin{figure}
\centerline{\psfig{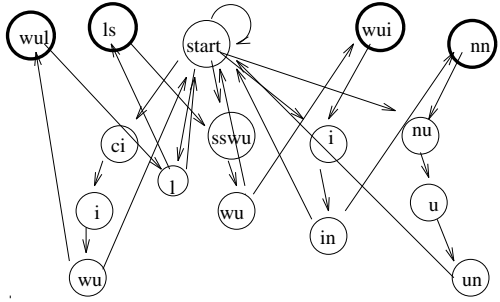}}
\caption{Trie-structured diphone-based HMM index for morphemes {\em ciwu, l, swu,
iss, nun}.  In each node, if a path from the root (start node) completes a morpheme, 
a pointer leads to the corresponding morpheme entry in the UPM dictionary.
The self-transition for each node is left out except the root for figure simplicity.}
\label{fg:header}
\end{figure}

The central part of the morphological analysis lies in the dictionary
construction. In our UPM (unified phonetic-morpheme) dictionary, each phonetic transcription of single 
morpheme
has a separate dictionary entry. Figure~\ref{fg:dict} shows the UPM
dictionary both for speech and language processing with three different morpheme 
entries {\em ci-wu, l, swu}. 
The extended morphological analysis is based on the well-known tabular 
parsing technique for context-free languages \cite{aho:theory} and augmented to 
handle the Korean
phonological rules and speech recognition errors in the diphone sequence 
inputs. Figure~\ref{fg:morph} shows 
the extended table-driven morphological analysis process. The example diphone
sequence was obtained from the input speech
{\it ci-wul-sswu} (meaning: can/cannot be removed), and the morphological 
analysis produces {\it ci-wu+l+swu} (remove+ADNOMINAL+BOUND-NOUN),
where '+' is the morpheme boundary, and '-' is the syllable boundary.
The morpheme segmentation is basically performed using the 
Viterbi-based lexical decoding to recover the possible errors in
the diphone sequences. For Viterbi search, the phonetic transcription
headers for each morpheme in the UPM dictionary are converted into diphone 
transcription headers, and each converted header is turned into a
simple HMM. The converted HMMs 
are organized into a trie data structure for efficient search (see
figure~\ref{fg:header}), and form a trie-structured diphone-based HMM index. The HMMs
are the simplest ones which have only left-to-right and self transitions.
Additional diphone nodes (marked with thick circles) are inserted for 
smooth inter-morpheme co-articulation modeling.  
The transition probability in each HMM is defined:
\[ a_{ij} = \left\{ \begin{array}{ll}
                     \alpha & \mbox{$i=j$} \\
                     \frac{1-\alpha}{N} & \mbox{$i \neq j \wedge d^{t} = s_{i}
\wedge d^{t+1} = s_{j}$} \\
                     0 & \mbox{otherwise}
                    \end{array}
            \right. \]
where $a_{ij}$ is a transition probability from state $i$ to state $j$, N
is the number of all possible transitions from state $i$. $d^{t}$ is a
diphone observable at time $t$, and $s_{i}$ is a diphone at state $i$. This
model assigns self-transition probability $\alpha$ and left-to-right
transition 
probability $\frac{1-\alpha}{N}$. All other transition probabilities are
zeros. In each state, the diphone emission probabilities are defined:
\[ b_{i} (k) = \left\{ \begin{array}{ll}
                       \beta & \mbox{$d_{k} = s_{i}$} \\
                       \frac{1-\beta}{M} & \mbox{otherwise}
                       \end{array}
               \right. \]
where $b_{i}(k)$ is a probability of producing diphone $d_{k}$ at state
$i$, and M is the number of all the diphones in the model. We
adjust $\alpha$ and $\beta$ experimentally, and the flexible adjustment helps
to cope with the insertion and deletion errors in the diphone sequences.
The Viterbi search with the trie-structured HMM index on  
the input diphone sequences segments out all the
possible morphemes in the given diphone sequence, and enrolls all the segmented
morphemes into the triangular table on the proper positions. For example, 
in figure~\ref{fg:morph}, morphemes such as {\em ci (carry), ci-wu
(delete), l (adnominal verb-ending), wul (cry), swu (bound-noun)} are
segmented out and enrolled in the table position (1,2), (1,3), (4,4), (3,4), 
(5,6). The position (i,j) designates the starting and ending position of
each morpheme in the given input {\em eonjeol}. 

The morphotactics modeling is necessary after all the morphemes are 
enrolled in
the table in order to combine only legal morphemes into an {\em eojeol} 
(Korean word), and the process is called morpheme-connectivity-checking. 
Since Korean has well developed postpositions (noun-endings, verb-endings,
prefinal verb-endings) which play as grammatical functional morphemes, we must
assign each morpheme proper part-of-speech (POS) tags for the efficient 
connectivity checking. Our more than 400 POS tags which are refined from 
the 13 major Korean lexical categories are 
hierarchically organized, and contained in the UPM dictionary
(in the name of left and right morphological connectivity, see figure~\ref{fg:dict}). 
In the case of idiomatic
expressions, we place such idioms directly in the dictionary for efficiency, 
where two different POS tags are necessary for the left and the right 
morphological connectivity. 
For single morpheme, the left and the right POS tags are always the same.
The separate morpheme-connectivity-matrix (sometimes, it is called
morpheme-adjacency-matrix) indicates
the legal morpheme combinations using the POS tags defined in the dictionary. 
So the morphotactics modeling
is performed by utilizing two essential components: the POS tags (in the dictionary) 
and the morpheme-connectivity-matrix. For example, in figure~\ref{fg:morph}, the
morpheme {\em ciwu} (in position (1,3)) can be legally combined with 
the morpheme {\em l} (in position (4,4)) to make {\em ciwu+l} (delete+ADNOMINAL, in
position (1,4)) but {\em ci} cannot be combined with {\em wul} to make 
{\em ci+wul} even if they are in the combinable positions. 

The orthographic rule modeling must be integrated with the phonological rule
modeling in spoken Korean processing. Since we must deal with the erroneous
speech
inputs, the conventional rule-based modeling requires so many number of
rule applications \cite{koskenniemi:two}. So our solution is based on the
declarative modeling of both orthographic and phonological rules in 
a uniform way.
That is, in our UPM dictionary, the conjugated verb forms as well as the original 
verb forms are all enrolled, and the same morphological connectivity 
information is applied for both original verb forms as well as the
conjugated ones.
The phonological rule modeling is also accomplished declaratively by having 
the separate phonemic
connectivity information in the dictionary (see figure~\ref{fg:dict}). The 
phonemic connectivity
information for each morpheme declares the possible phonemic changes 
in the first (left) and the last (right) positioned phonemes in the morpheme, 
and the 
phoneme-connectivity-matrix indicates the legal sound combinations in Korean
phonology using the defined phonemic connectivity information. For example, 
in figure~\ref{fg:morph}, the morpheme {\em l} can be
combined with the morpheme {\em swu} during the morpheme connectivity checking 
even if {\em swu} is actually pronounced as {\em sswu} (see the input in
figure~\ref{fg:morph}). The 
phoneme-connectivity-matrix supports the legality of the
combination of {\em l} sound with changed {\em s to ss} sound. This legality comes
from the Korean phonology rule {\em glotalization} (one form of consonant
dissimilation) stating that {\em s} sound becomes {\em ss} sound after 
{\em l} sound. In this way, we can declaratively model all the major Korean
phonology rules such as (syllable-final consonant) standardization, consonant 
assimilation,
palatalization, glotalization, insertion, deletion, and contraction. 

\section{Implementation and experiment results}
\label{sec:imple}
The SKOPE speech and natural language integration architecture was 
implemented using a standard 
C and X-window user interface on a UNIX/Sun Sparc 
platform. The system's inputs are carefully articulated Korean speeches
in a normal laboratory environment, and the outputs are morphologically
analyzed {\em eojeol} sequences which can be directly used by
Korean syntactic and semantic analysis modules.  
We constructed a 1000 morpheme-entry UPM dictionary in a UNIX operating system 
domain \cite{lee:from}, and built morpheme connectivity and phoneme connectivity 
matrices for the phonological/morphological co-analysis. The UPM dictionary is 
indexed using the diphone transcribed HMM headers for each morpheme,
which are organized into a trie. Since we don't have any
standard segmented Korean speech database yet, we constructed our
own by recording and manually segmenting 73 most frequent Korean diphones.
The 73 diphones are acquired from the 300 Korean {\em eojeols} 
(each {\em eojeol} is pronounced 15 times by a female speaker) in 50 Korean 
sentences which can appear in natural language  
commanding to the UNIX operating system \cite{lee:from}. 

Several experiments were performed to verify the system's performance
of time-shift invariance, diphone recognition, and final {\em eojeol} 
recognition including the spoken language morphological analysis.
In each experiment, the input speech patterns were prepared as follows:
{\em eojeols} were recorded in a normal laboratory environment with an average
S/N ratio of 12 dB. Speech data were sampled at 16kHz-16bit, and hamming-windowed.
From this windowed data, 512-point DTFTs were computed at 5 msec intervals.
The DTFTs were used to generate 16 Mel-scale filter-bank coefficients at
10 msec frame size \cite{waibel:phoneme}. These spectra were normalized to
produce suitable input levels for the four-layer TDNNs.
We used hyperbolic arc tangent error function for the weight
updating \cite{fahlman:fast} in the back propagation training, and updated
the weights after a small number of iterations \cite{haffner:fast}.

\subsection{Time-shift invariance of Korean diphones}
\begin{figure}
\centerline{\psfig{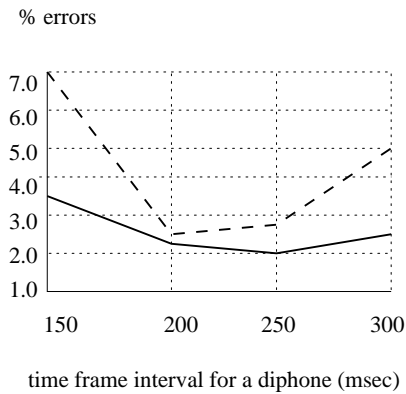}}
\caption{Average error rate of the segmented time frame (solid lines) versus
the same time frame with maximum 40 msec left or right temporal shift (dotted
lines)}
\label{fg:titest}
\end{figure}

We generated 2400 diphone samples for typical 12 Korean diphones.
The input patterns for two test cases are set the same in order to 
compare the {\em no-time-shift} and {\em time-shift} cases. 
Figure~\ref{fg:titest} shows that the
Korean diphone recognition maintains the time shift invariance property of TDNN and
suggests the optimal time interval near 200 - 250 msec. 

\subsection{Comparison of diphone recognition vs. phoneme recognition}
\begin{figure}
\centerline{\psfig{figure=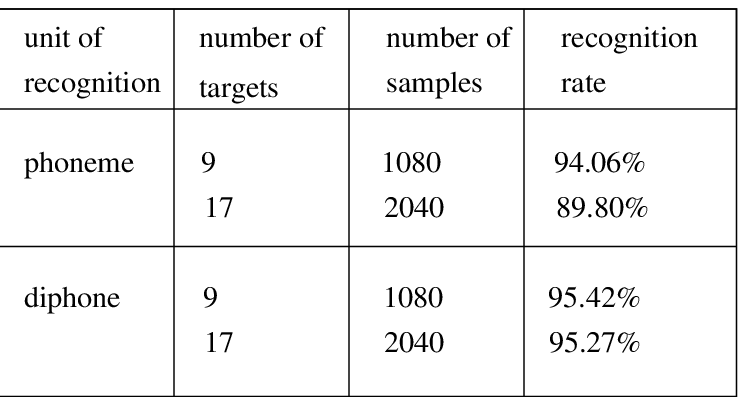}}
\caption{Diphone recognition versus phoneme recognition test}
\label{fg:di-perf}
\end{figure}

This experiment is to show that diphones can improve the recognition
rate of Korean vowels regardless of many rising diphthongs
compared with the phoneme recognition. In the test, we set
150 msec time range for the phoneme and 200 msec for the diphone
segmentation. Compared with  
the phoneme recognition, figure~\ref{fg:di-perf} shows that diphone
recognition performance doesn't drop much when the number of targets
with similar features doubly increases. 

\subsection{Performance of continuous diphone recognition}
\label{sc:pdi}
\begin{figure}
\centerline{\psfig{figure=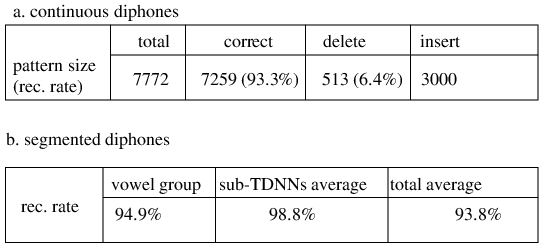}}
\caption{Continuous diphone spotting versus segmented diphone recognition}
\label{fg:cdi-perf}
\end{figure}

In this experiment, we pronounced carefully chosen 66 {\em eojeols} 15 times 
to generate about 5500 diphone patterns for training. The 5500
training samples are used to train the vowel group TDNN and 18 different
sub-TDNNs for each diphone group. During the recognition, the new 262
eojeols are selected to generate the test patterns of 2432 eojeols, and
these test patterns are shifted 30 msec during the recognition to obtain the 
TDNN diphone 
spotting performance in a continuous speech. Figure~\ref{fg:cdi-perf}-a shows
the continuous diphone spotting performance. 
We have total 7772 target diphones from the 2432 test eojeol patterns.
The {\em correct} designates that the correct
target diphones were spotted in the testing position, and the {\em delete} 
designates the other case (including the substitution errors).
The {\em insert} designates that the non-target diphones were spotted in the 
testing position. To compare the ability of handling the continuous speech,
we also tested the diphone spotting using the hand segmented test patterns
with the same 7772 target diphones.
Figure~\ref{fg:cdi-perf}-b shows the segmented diphone recognition performance.
Since the test data are already hand-segmented before input, there are no
insertion and deletion errors in this case.
The fact that the segmented speech performance is not much better than the
continuous one (93.8\% vs. 93.3\%) demonstrates the 
diphone's suitability to handling the Korean continuous speech. 

\subsection{Performance of continuous speech morphological analysis}
\begin{figure}
\centerline{\psfig{figure=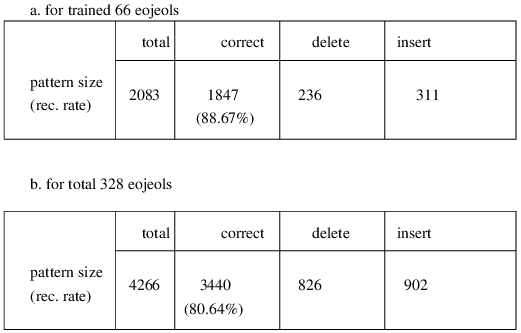}}
\caption{Continuous speech input morphological analysis performance}
\label{fg:cej-perf}
\end{figure}

In order to test the ability of full {\em eojeol} recognition 
including the Viterbi-based lexical decoding and 
phonological/morphological co-analysis,
a middle-vocabulary experiment was carried out. The task is
a speaker-dependent and continuous eojeol recognition which produces the
morphologically analyzed eojeol sequences directly from the speech inputs. 
In the process, the speech recognizer produces the erroneous diphone sequences  
in input {\em eonjeols}, and then  
the Viterbi morphological analyzer segments them with the error correction and produces
the final analyzed {eojeols}. So, in this task, all the intermediate steps, that
is, diphone spotting, lexical decoding and morphological/phonological analysis, 
are combined to produce the final recognition
performance.  The same 328 eojeols in section~\ref{sc:pdi} were fed to the 
SKOPE integration architecture that has the pre-trained TDNNs (with 66 eojeols).
Figure~\ref{fg:cej-perf}-a shows the performance with the trained 66
{eojeols} and figure~\ref{fg:cej-perf}-b shows the final performance of 
the total 328 eojeols. We have total 4266 target morphemes from the same 
328 eojeols used in section~\ref{sc:pdi}.
In the figure, the {\em correct} designates that the correct morpheme
sequences can be analyzed from the speech input, and the {\em delete} means
that the correct morpheme sequences cannot be generated (including the
substitution errors). The {\em insert}
designates the percentage of the spurious morphemes that are generated from 
the insertion errors. 
The performance is about 80.6\% correctness in the final morphological analysis
with the mostly untrained new data, which is quite promising considering the
complexity of the task.

\section{Comparison with related researches}
\label{sec:comp}
Recently, the idea of sending only n best speech recognition 
results to a natural language module has been implemented using the
time-synchronous Viterbi-style search algorithm \cite{chow:nbest}. The
algorithm was also improved by the word-dependent search 
\cite{schwartz:efficient} and by adding the A* backward tree 
search \cite{soong:tree}. The n-best
integration scheme has been mostly utilized for HMM-based continuous 
speech recognition systems, and many existing speech systems and 
natural language 
systems were successfully integrated using the n-best word search techniques
\cite{agnas:spoken,bates:bbn}.  However, until now, the n-best search 
techniques are
only implemented to directly produce the n-best sentences using the word
sequences, and this word-level integration is inefficient for
the morphologically complex languages such as Korean. On the contrary, our
integration is at the morpheme-level directly decoding the PLU sequences 
with the morphological processing because we need more sophisticated 
phonological/morphological handling in the
early stage of the integration process. The word-level n-best integration 
also assumes
the word-level dictionary which is an unreasonable assumption for 
morphologically complex languages. According to the Harper and others'
recent classification \cite{harper:integrating}, n-best integration is a 
typical loosely-coupled example.  

The HMM-LR integration \cite{kita:hmm,hanazawa:atr} was implemented 
using the HMM's phoneme spotting ability integrated with 
the generalized LR parsing
techniques \cite{tomita:efficient}. Unlike the n-best integration, the
HMM-LR integration was more tight and implemented at the phoneme-level 
by extending the LR
parser's terminal symbols to cover the phonetic transcriptions. 
In this scheme, the LR parsing 
selects the most probable parsing results by obtaining the probability of 
the end-point candidate phonemes from the HMM's forward probability 
calculation. So
the total integrated system is working by the LR parser's prediction 
of the next phoneme candidates which are then verified by the HMM's phoneme 
spotting abilities.  The idea of extending the LR grammar to the phonetic
transcriptions seems to be working for the phoneme-level integration.
However, the scheme doesn't have any separate language-level dictionary, 
which results in the degenerated phonological/morphological processing, and
also suffers from difficulty in the necessary scale-ups. On the contrary, our SKOPE 
integration architecture focuses on the general phonological/morphological 
handling during the integration which is essential for the agglutinative 
languages.
The idea of extending LR grammar to the phonetic transcriptions was also 
applied to the TDNN-LR integration method \cite{sawai:tdnn-lr,sawai:the}
which was similarly implemented by replacing HMM's phoneme spotting by the
TDNN's phoneme spotting. The integration was implemented by dynamic time
warping (DTW) level-building search \cite{myers:level} between TDNN's
phoneme sequences and LR grammar's phoneme sequences. However, the
performance was relatively poor compared with the HMM-LR integration method
\cite{sawai:tdnn-lr}.   
There are basically two reasons for the poor TDNN-LR performance compared
with the HMM-LR integration:
1) the TDNN model has rarely been applied to the practical large
vocabulary systems yet, therefore it lacks the fine tuning compared with
the popular HMM models, and 2) the TDNN model has yet to find a
right way to be effectively integrated into the natural language processing
model. The HMM model supports a natural integration into the
general chart-based parsing models such as generalized LR parsing 
because there are well-defined probablistic search techniques 
in the language as well as in the speech levels. 
However, output activations of the multiple TDNNs are difficult to normalize
and therefore difficult to be naturally integrated 
into the popular probabilistic search schemes such as
Viterbi search. Our SKOPE architecture adopts Viterbi search with
pre-defined transition and emission probabilities, and use the Viterbi
search for only segmenting erroneous diphone strings. All the other
morphological processing steps are generally performed according to the 
symbolic natural language processing model. 

The more tightly-coupled systems have also been researched to integrate all 
the knowledge sources of spoken
language processing from acoustic to semantic into a single interdependent
model that cannot easily be separated. In these systems, syntactic parser
directly deals with acoustic-level inputs. For example, Ney
\cite{ney:dynamic} extended CYK parsing algorithm to cover acoustic inputs
by exhaustively finding all possible endpoints for every terminal symbol. 
In the similar vein, the HMM can be extended to handle recursive embedding
for context-free grammar processing \cite{lari:applications}. However, these
acoustic-level syntactic parsers are computationally expensive since the
parsing complexity is at best $O(n^{3})$ where n could be in several hundreds
when the parsers directly deal with the speech frames. The SKOPE integration is tighter 
than loosely-coupled 
n-best techniques, but less tight compared with these tightly-coupled systems. 
We agree that the high-level linguistic constraints should restrict the underlying
speech recognition in some ways as in the tightly-coupled systems, but disagree 
that the constraints should be in a syntax level. The
more tightly-coupled systems are often impractical for
large-scale spoken language processing because of the time complexity. Moreover, we
still don't have much knowledge to tell how much top-down feedback is actually
helpful to improve the speech recognition process. As an engineering point of
view, semi-tightly-coupled systems are quite feasible for large complex systems
under the current technology. In this regard, SKOPE project adopts
a semi-tightly-coupled integration technique between speech and language processing, especially
morphological processing. 

\section{Conclusions}
\label{sec:con}
This paper presents a morpheme-level integration architecture of speech 
and natural language in a connectionist continuous speech recognition model 
for agglutinative languages such as Korean. Our
main contributions are to present the morphologically conditioned semi-tight 
integration model that can support sophisticated phonological/morphological
processing in the integration of speech and language, 
which is essential for morphologically complex agglutinative languages.  
Also, the SKOPE integration architecture is a first attempt to develop a
morphologically general integration model using the connectionist speech
recognition paradigm.  

The SKOPE speech and language integration architecture has many novel
features for speech and natural language processing.  First, the
diphone-based TDNN proposes a nice sub-word unit of recognition,
well reflecting the Korean phonetic characteristics. Secondly, 
the morphological analysis combined with the declarative phonological rule 
modeling is well suited
to the phonetic spelling into the orthographic morpheme mapping, which is 
an essential task for every spoken language processing model. 
Finally, the trie-structured HMM indexing for UPM dictionary
enables the Viterbi style search to be applied to the thorny morpheme
segmentation and lexical decoding problem, and also provides natural
integration of symbolic natural language processing techniques with 
probabilistic decoding schemes. 
The experiments show that the final morphological analysis performance from 
continuous speech is over 80.6\% in a
middle-vocabulary speaker-dependent recognition task, 
which is very promising in
considering the continuous speech and the combination of several steps of
performances such as diphone spotting, lexical decoding and
morphological/phonological analysis. 
Since the integration architecture is based on general linguistic notion of 
phoneme and morpheme, the architecture is not restricted to Korean.  
The SKOPE architecture can be extended to any agglutinative language which has
clear-cut morphological boundaries such as Japanese, and possibly to other
Indo-European languages which exhibit well-developed morphological phenomena such as
German. We are now extending the integration technique to Japanese. 

\section*{Acknowledgements}
This research was partly supported by grants from KOSEF
and ETRI in Korea. Many of our former or current students implemented
the idea and we thank all of them. Especially, we thank Kyunghee Kim for
implementing diphone-based Korean speech recognizer,
WonIl Lee for coding the lexicon and the morphological
parser, and finally ByungChang Kim for implementing Viterbi search for the
diphones. Professor Hong Jeong directed us to the deep world of signal
processing for the earlier part of the project.

\bibliographystyle{ieeetr}

\end{document}